\documentclass[conference]{IEEEtran}

\usepackage[nocompress]{cite}
\usepackage[utf8]{inputenc}
\usepackage{amsmath}
\usepackage{hyperref}
\usepackage{graphicx}
\usepackage{caption}
\usepackage{subcaption}
\usepackage[inline]{enumitem}
\usepackage{amsmath}
\usepackage{amssymb}

\DeclareMathOperator*{\argmax}{arg\,max}

\usepackage{tikz}
\usetikzlibrary{trees}

\title{Unifying Evaluation of Machine Learning Safety Monitors}

\author{\IEEEauthorblockN{Joris Guerin}
\IEEEauthorblockA{\textit{Univ. Toulouse, LAAS-CNRS}\\
Toulouse, France \\
jorisguerin.research@gmail.com}
\and
\IEEEauthorblockN{Raul Sena Ferreira}
\IEEEauthorblockA{\textit{LAAS-CNRS}\\
Toulouse, France \\
rsenaferre@laas.fr}
\and
\IEEEauthorblockN{Kevin Delmas}
\IEEEauthorblockA{\textit{ONERA}\\
Toulouse, France \\
kevin.delmas@onera.fr}
\and
\IEEEauthorblockN{Jérémie Guiochet}
\IEEEauthorblockA{\textit{Univ. Toulouse, LAAS-CNRS}\\
Toulouse, France \\
jeremie.guiochet@laas.fr}
}

\begin{document}

\maketitle


\begin{abstract}
With the increasing use of Machine Learning (ML) in critical autonomous systems, runtime monitors have been developed to detect prediction errors and keep the system in a safe state during operations. Monitors have been proposed for different applications involving diverse perception tasks and ML models, and specific evaluation procedures and metrics are used for different contexts. This paper introduces three unified safety-oriented metrics, representing the safety benefits of the monitor (\emph{Safety Gain}), the remaining safety gaps after using it (\emph{Residual Hazard}), and its negative impact on the system's performance (\emph{Availability Cost}). To compute these metrics, one requires to define two return functions, representing how a given ML prediction will impact expected future rewards and hazards. Three use-cases (classification, drone landing, and autonomous driving) are used to demonstrate how metrics from the literature can be expressed in terms of the proposed metrics. Experimental results on these examples show how different evaluation choices impact the perceived performance of a monitor. As our formalism requires us to formulate explicit safety assumptions, it allows us to ensure that the evaluation conducted matches the high-level system requirements.
\end{abstract}


\section{Introduction}\label{sec:intro}
Recent breakthroughs in Machine Learning (ML) slowly allow autonomous systems to operate in the real world, where failures can be catastrophic, e.g., self-driving cars~\cite{calvi2019runtime}. This work focuses on ML-based perception functions that interpret complex sensor signals to estimate state~\cite{premebida2018intelligent}, e.g., pedestrian detection~\cite{brunetti2018computer}, for which there is currently no valid alternative to complex ML models. However, the use of ML has raised new dependability challenges such as the lack of well-defined specification, the black-box nature of the models, the data high-dimensionality, and the over-confidence of neural networks~\cite{faria2018machine,mohseni2019practical}. Consequently, offline actions are not sufficient to guarantee the safety of such critical autonomous systems. As an alternative, recent research investigated online fault tolerance mechanisms, which we refer to as runtime monitors, to maintain an acceptable behavior during operation despite perception errors. 

\begin{figure*}[!ht]
    \centering
    \includegraphics[width=0.8\textwidth]{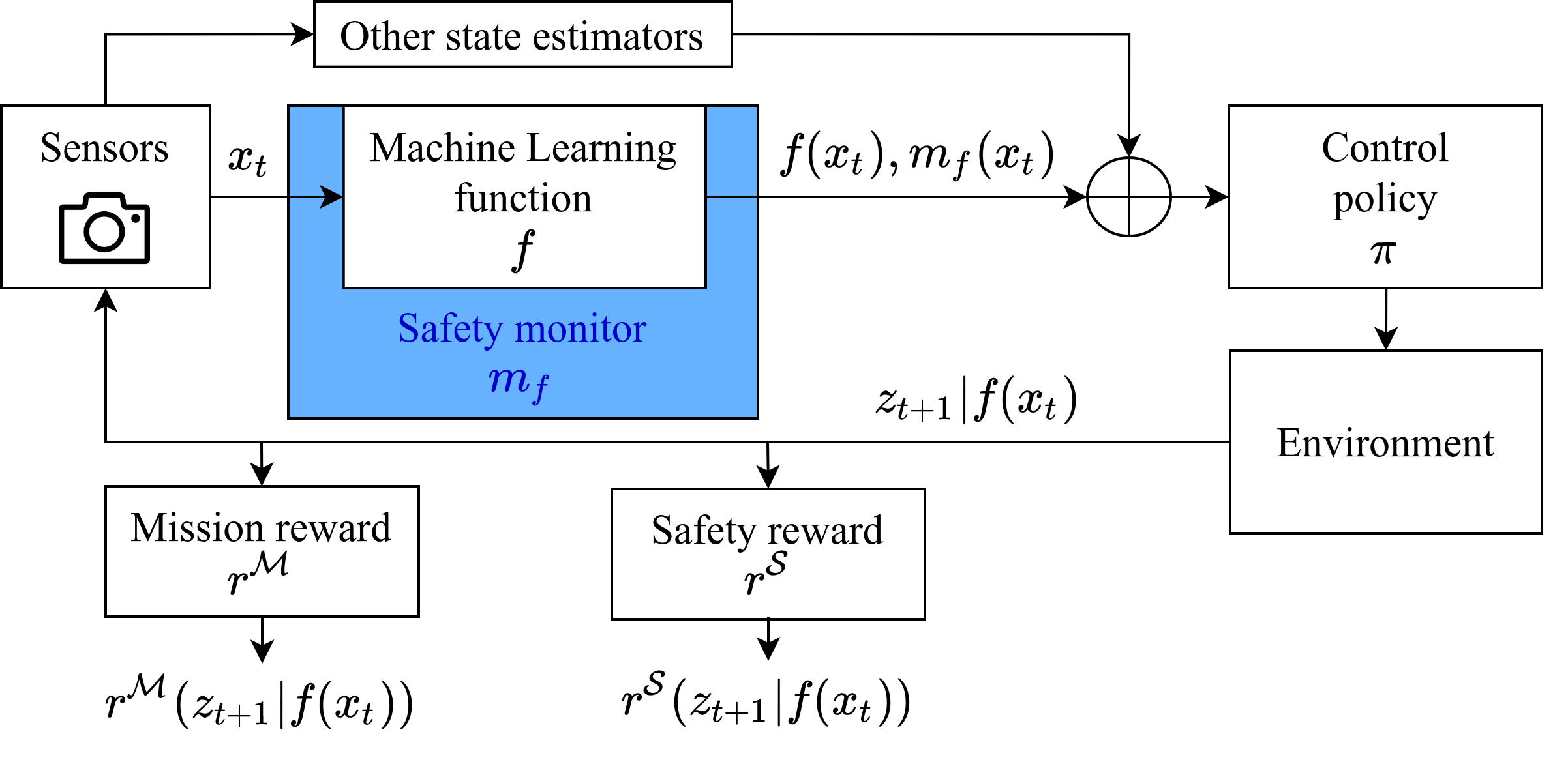}
    \caption{\textbf{Machine Learning-based system.} This representation is well suited to describe an autonomous system relying on ML for state estimation. The ML function can be enhanced with a runtime safety monitor to detect hazardous errors.}
    \label{fig:system}
\end{figure*}

Runtime monitors are safety components acting close to the perception function of interest, in charge of detecting hazardous errors and raising alert to the system accordingly~\cite{rahman2021run} (\autoref{fig:system}). A good monitor should increase the system's \emph{safety} (absence of hazardous situations) without decreasing its \emph{availability} (ability to perform its mission). Recent works have proposed to develop specific runtime monitors for a variety of visual perception tasks (e.g., classification~\cite{hendrycks2016baseline,doctor,dissector}, object detection~\cite{carla_OD}, semantic segmentation~\cite{icra22}, steering angle regression~\cite{e2e_AD2,taxiing}). Depending on the application context, distinct evaluation procedures and metrics have been used to quantify the performance of runtime monitors. Such evaluation choices actually reflect safety assumptions about the system of interest, which are rarely explicit. For example, the most common assumption for classification tasks is that any misclassification has the same impact on the system's safety (see \autoref{sec:examples_classif}). This paper aims to unify the evaluation methodology for runtime monitors across tasks and application contexts by defining evaluation metrics that:
\begin{enumerate*}
    \item capture the safety benefits of the evaluated monitor,
    \item capture the remaining safety gaps despite using the monitor,
    \item capture the negative impact of the monitor on system's availability,
\end{enumerate*}

Evaluation metrics for generic non-ML runtime monitors (or checkers) were discussed in~\cite{checkers_eval}. The concept of checker coverage was introduced to represent the probability of failure of a primary-checker fault-tolerant architecture. However, in their formulation, they assume that an output of the monitored component is either a success or a failure and that this binary status fully determines the resulting hazard. The output can be partially correct for ML-based perception functions, e.g., a pedestrian detection model locates only a subset of individuals in an image. In addition, all errors from an ML model do not lead to equally hazardous situations, e.g., an autonomous vehicle not detecting a pedestrian on the sidewalk. Our proposed metrics are sufficiently flexible to model these specificities of ML-based functions.

An attempt of generic formulation for ML functions was proposed among the \textit{selective prediction} community~\cite{geifman2019selectivenet}. Their evaluations introduce the notions of coverage and selective risk, representing the size of the region of the input space where the monitor does not activate and the hazard associated with this region. By introducing a customizable loss function to represent hazard, this formulation is well suited to define the safety impact of a wrong acceptance. However, unlike our approach, these evaluation metrics lack the flexibility to model the potential negative effects of wrong rejections on safety (relevant example in \autoref{sec:examples_EL}) and system availability.

To address the above limitations, in Section~\ref{sec:approach}, we 
introduce new generic metrics called \emph{Safety Gain}, \emph{Availability Cost} and \emph{Residual Hazard} that present the desired properties. These metrics consist in estimating future cumulative safety and mission rewards following a given prediction.
Then, in Section~\ref{sec:examples}, we demonstrate in three distinct use-cases that diverse evaluation metrics from the literature can easily be expressed in terms of the proposed metrics. These examples show that using our formalism helps to ensure that the evaluations conducted are in line with the actual safety requirements of the system. Finally, in Section~\ref{sec:conclusion}, we discuss the benefits and limitations of this work.

\section{Proposed evaluation formalism}\label{sec:approach}


This section presents unified evaluation metrics for runtime safety monitors in their generic form. Relaxations are also introduced to make them usable in practice. Note that practical examples of usage are presented in Section~\ref{sec:examples}.

\subsection{Context and notations}

Let $f$ be a predictive model, in charge of approximating an unknown function $f^*$ over an operational design domain $\mathcal{D}$\footnote{Set of all situations under which the system is expected to work, e.g., for autonomous vehicles it can be described by roadway types, geographic characteristics, speed ranges, and weather conditions, among others~\cite{ODD_AV}.}. In particular, this paper focuses on complex perception functions used in critical autonomous systems to estimate the state of the agent and/or its environment, e.g., pedestrian detection. In other words, the system uses the outputs of $f$ to decide what actions to take in the real world, and these actions can impact the safety of the agent and its surroundings. Figure~\ref{fig:system} presents a simplified generic architecture for such a system, which can be used to better understand our notations. A set of initial conditions, fully defining the state of the system and its environment in a given world, is called a scenario. In this work, we consider that the behavior of the autonomous system is optimal if $f$ works perfectly, i.e., if for all possible scenarios and time steps~$t$, $f$ processes the sensor input $x_t$ as expected ($f(x_t) = f^*(x_t)$), then the actions performed are safe and allow to fulfill the system's mission. 

However, for such perception tasks, $f$ is often built using complex ML models such as deep neural networks, which are known to make errors that are difficult to predict. Let $z_{t+1}|f(x_t)$ be the world state (system + environment) at time step $t+1$, when the autonomous system follows its policy $\pi$ after receiving a prediction $f(x_t)$ from the perception model. To simplify notations, we assume that $x_t$ is representative of the world state $z_t$, and that the system satisfies the Markov property, i.e., the environment's response at $t+1$ depends only on the state and action at~$t$. 
Let~$r^{\mathcal{M}}(z)$ be the \emph{mission reward}, which associates a score to a configuration~$z$, representing the progress of the system with respect to its mission. Likewise, let~$r^{\mathcal{S}}(z)$ be the \emph{safety reward}, representing the safety of configuration~$z$. In practice, the function~$r^{\mathcal{M}}$ is defined during the design phase, and the function~$r^{\mathcal{S}}$ results from the safety analysis of the system (the safety score is inversely proportional to the residual hazards of a configuration). For example, if an autonomous vehicle is running normally on the road at time step~$t$, $r^{\mathcal{S}}(z_t)$ should be high, but if it collides with a pedestrian at time step~$t'$, $r^{\mathcal{S}}(z_{t'})$ should be very low. Some concrete examples of how~$r^{\mathcal{S}}$ can be defined are presented in Section~\ref{sec:examples}.


Then, we define the \emph{mission return} associated with $f$ after a given prediction at time step~$t$:
\begin{equation}\label{eq:defReturnMission}
    R^{\mathcal{M}}_f(x_t) = \sum_{k=t+1}^{T} r^{\mathcal{M}}(z_k|f(x_t)),
\end{equation}
where $T$ is the length of the episode\footnote{In this paper, the formalism is presented in the finite-horizon setting, but extension to infinite-horizon should be straightforward.}.
$R^{\mathcal{M}}_f(x_t)$ represents the cumulative mission reward resulting from $f(x_t)$. Likewise, the \emph{safety return} is defined as:
\begin{equation}\label{eq:defReturnSafety}
    R^{\mathcal{S}}_f(x_t) = \sum_{k=t+1}^{T} r^{\mathcal{S}}(z_k|f(x_t)).
\end{equation}
A low~$R^{\mathcal{M}}_f(x_t)$ means that the prediction~$f(x_t)$ will decrease the availability of the system, and a low~$R^{\mathcal{S}}_f(x_t)$ means that~$f(x_t)$ may lead to an unsafe state. Under these notations, we assume that the unknown function~$f^*$ always makes the best possible predictions, that is, $\forall t$: 
\begin{equation*}
\begin{array}{l}
f^* \;\; = \;\; \argmax_{f} \; R^{\mathcal{M}}_f(x_t), \\ \\
f^* \;\; = \;\; \argmax_{f} \; R^{\mathcal{S}}_f(x_t).
\end{array}
\end{equation*}
We acknowledge that the sequence of consecutive states $\{z_{t+1}|f(x_t), ..., z_{T}|f(x_t)\}$ is not known a priori and is most likely non-deterministic. However, we will see later how this formulation can be used in practice to evaluate the impact of the predictions of~$f$.

\subsection{General formulation of the proposed metrics}
As it is hard to guarantee that~$f$ is valid across the entire operational design domain~$\mathcal{D}$, it is essential to equip such complex ML perception functions with appropriate runtime monitoring mechanisms to detect errors of~$f$ and maintain the system in a safe state. Let~$m_{f}$ be such a monitor for~$f$, i.e., for a given input~$x\in \mathcal{D}$, $m_f$ should detect when~$f(x)$ will be erroneous and raise an alert accordingly. The remaining of the system can then modify its actions, thus defining a new forward state sequence $\{z_{t+1}|(f,m_f)(x_t), ..., z_{T}|(f,m_f)(x_t)\}$.
This way, $R^{\mathcal{M}}_{(f, m_f)}(x_t)$ and $R^{\mathcal{S}}_{(f, m_f)}(x_t)$ depend not only on the values returned by $f(x_t)$, but also on the status of the monitor $m_f(x_t)$. In this paper, we only consider binary monitors that activate and raise alerts when they judge that $f(x_t)$ is hazardous for the system ($m_f(x_t) = 1$), and do nothing otherwise ($m_f(x_t) = 0$). 

When evaluating the performance of a monitor $m_{f}$, we are interested in quantifying the three following measures detailed hereafter:
\begin{description}
    \item[$SG_{m_f}$:] The safety improvements resulting from $m_f$. 
    \item[$RH_{m_f}$:] The remaining hazard in the system after using $m_f$.
    \item[$AC_{m_f}$:] The system performance decrease because of $m_f$.
\end{description}

\subsubsection{Safety Gain} To know the overall safety added by~$m_f$ across the operational design domain $\mathcal{D}$,  one needs to compare the safety of the monitored system $(f, m_f)$ against the safety of the initial system $f$. This objective is captured by the Safety Gain metric defined as:
\begin{equation}\label{eq:SG}
        SG_{m_f} = \int_{\mathcal{D}} p(x)\left(R^{\mathcal{S}}_{(f, m_f)}(x) - R^{\mathcal{S}}_{f}(x)\right) \, dx,
\end{equation}
where $p(x)$ represents the likelihood of $x$ when randomly sampling $\mathcal{D}$. $SG_{m_f}$ represents the ``amount of hazard'' that was removed from the system by implementing $m_f$. Integrating over $\mathcal{D}$ corresponds to averaging across all possible scenarios, noting that a state $z_{t\neq 0}$ for be used as initial conditions to define another scenario. A good monitor should have $SG_{m_f}$ as high as possible. If the safety scores of the different threats identified during the preliminary safety analysis of the system are scaled in $[0,1]$, then $SG_{m_f}$ ranges between $-1$ and $1$. If it is not the case, we can still scale it by dividing the results by $\int_{\mathcal{D}} (r^{\mathcal{S}}_{\text{max}}) \, dx$, where $r^{\mathcal{S}}_{\text{max}}$ is the maximum possible safety reward of an event.

\subsubsection{Residual Hazard} To model the hazard still present in the system despite $m_f$, one needs to compare the safety of the monitored system $(f, m_f)$ against the safety of the optimal model $f^*$. This objective is captured by the Residual Hazard metric defined as:
\begin{equation}\label{eq:RH}
        RH_{m_f} = \int_{\mathcal{D}} p(x) \left(R^{\mathcal{S}}_{f^*}(x) - R^{\mathcal{S}}_{(f,m_f)}(x)\right) \, dx.
\end{equation}
It represents the ``amount of hazard'' that is still present in the system, despite implementing $m_f$. A good monitor should have $RH_{m_f}$ as low as possible. Regarding scaling of $RH_{m_f}$, considerations defined above are also valid. 

\subsubsection{Availability Cost} 
\begin{figure}[t]
    \centering
    \includegraphics[width=0.48\textwidth]{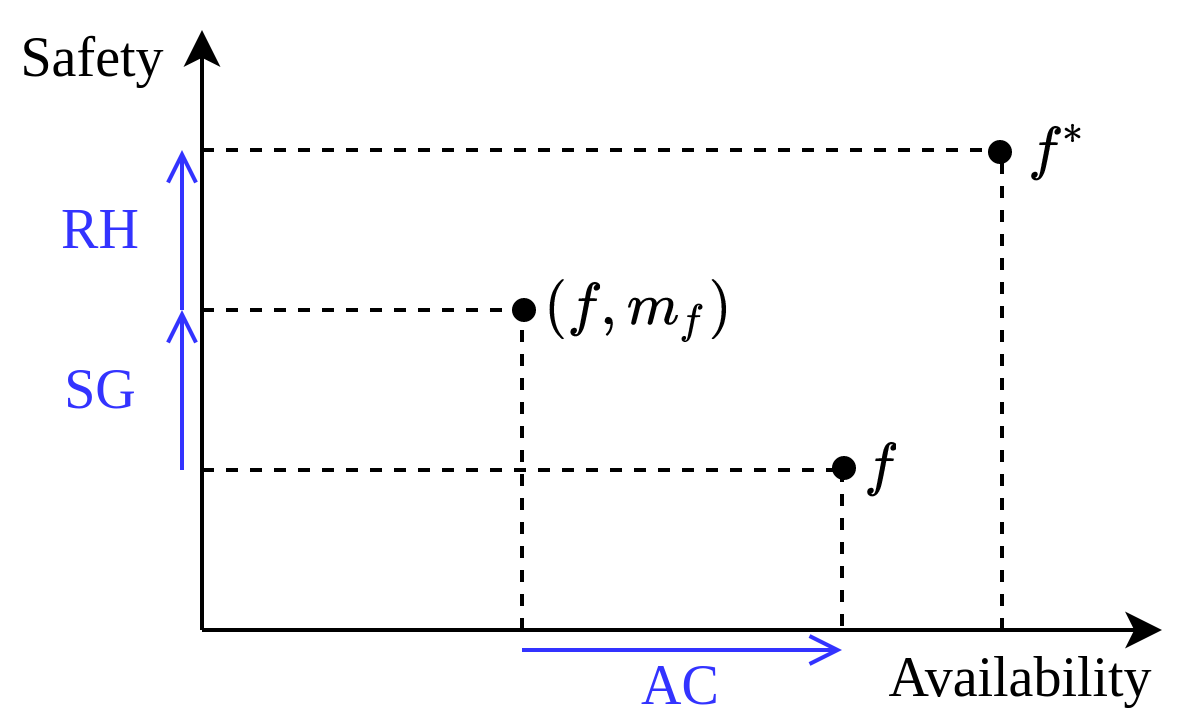}
    \caption{\textbf{Representation of the proposed metrics}. The Safety Gain (SG), Residual Hazard (RH) and Availability Costs (AC) are computed with respect to the returns of the predictive function $f$, its runtime monitor $m_f$ and the ground truth function $f^*$.}
    \label{fig:metrics_diag}
\end{figure}

To model the decrease in system's performance due to $m_f$ across $\mathcal{D}$, one needs to compare the availability of the monitored system $(f, m_f)$ against the availability of the initial system $f$. This objective is modeled by the Availability Cost metric defined as:
\begin{equation}\label{eq:AC}
        AC_{m_f} = \int_{\mathcal{D}} p(x) \left(R^{\mathcal{M}}_{f}(x) - R^{\mathcal{M}}_{(f, m_f)}(x)\right) \, dx.
\end{equation}
It represents the ``amount of mission reward'' that was lost by implementing $m_f$. A good monitor should have $AC_{m_f}$ as low as possible. Regarding scaling of $AC_{m_f}$, similar considerations apply. A visual representation of SG, AC and RH can be seen in \autoref{fig:metrics_diag}.

We also note that $AC_{m_f}$ and $RH_{m_f}$ are complementary. One represents the amount of existing hazard that was removed by $m_f$, while the other is the amount that was not handled by $m_f$. Their sum represents the hazard associated with $f$:
\begin{equation}
    SG_{m_f} + RH_{m_f} = \int_{\mathcal{D}} p(x) \left(R^{\mathcal{S}}_{f^*}(x) - R^{\mathcal{S}}_{f}(x)\right) \, dx.
\end{equation}

\subsection{Relaxations}\label{sec:approach_relax}

For most practical applications, the metrics defined above cannot be computed. In this section, we will explain why and propose relaxations of these metrics to make them usable in practice. Practical examples to show how diverse evaluation approaches from the literature can be expressed using these relaxations are presented in Section~\ref{sec:examples}.

The first issue is most of the relevant problems are too complex to derive an exact model of the behavior of the system and its entire environment under all possible external conditions (weather, illumination, etc.). This is particularly true for autonomous systems with diverse long-term behavior. For a given $x \in \mathcal{D}$, the subsequent states cannot be computed precisely, which makes $R^{\mathcal{M}}_{(f,m_f)}(x)$ and $R^{\mathcal{S}}_{(f,m_f)}(x)$ almost impossible to compute. A solution to this problem is to develop an alternative way to approximate these returns using whatever information is available. We note such approximations $\hat{R}^{\mathcal{M}}_{(f,m_f)}(x)$ and $\hat{R}^{\mathcal{S}}_{(f,m_f)}(x)$. In Section~\ref{sec:examples}, we show how very different evaluation approaches from the literature can be presented simply as procedures to compute the safety and mission returns.

The second issue is that the boundaries of the operational design domain $\mathcal{D}$, and by extension, the optimal function $f^*$ are unknown a priori. This lack of specification of the perception function is precisely why ML was used in the first place. Without formal knowledge of $\mathcal{D}$ and $f^*$, it is impossible to compute any of the metrics defined in Equations~\ref{eq:SG}, \ref{eq:RH} and \ref{eq:AC}. To address this limitation, one can conduct an evaluation on a predefined dataset, where true values of $f^*$ are available for a specific subset of points in $\mathcal{D}$. In other words, the evaluation is conducted on a labeled dataset $D_{\text{eval}} = \{(x_1, y_1), ..., (x_n, y_n)\}$ such that $\forall i \in \{1, ..., n\}, \; x_i \in \mathcal{D}, \; y_i = f^*(x_i)$. The underlying assumption for this relaxation is that $D_{\text{eval}}$ is representative of the operational design domain, i.e., it is large enough and was sampled from the expected application context (distribution of $\mathcal{D}$). Characterizing the test coverage of a specific dataset is an active field of research~\cite{test_coverage}, but it is beyond the scope of this paper.


Combining the two relaxations above, one can then proceed to compute estimates of the three evaluation metrics:
\begin{equation}\label{eq:SG_approx}
        \hat{SG}^{D_\text{eval}}_{m_f} = \frac{1}{n} \sum_{i=1}^n \left(\hat{R}^{\mathcal{S}}_{(f, m_f)}(x_i) - \hat{R}^{\mathcal{S}}_{f}(x_i)\right),
\end{equation}
\begin{equation}\label{eq:RH_approx}
        \hat{RH}^{D_\text{eval}}_{m_f} = \frac{1}{n} \sum_{i=1}^n \left(\hat{R}^{\mathcal{S}}_{f^*}(x_i) - \hat{R}^{\mathcal{S}}_{(f,m_f)}(x_i)\right),
\end{equation}
\begin{equation}\label{eq:AC_approx}
        \hat{AC}^{D_\text{eval}}_{m_f} = \frac{1}{n} \sum_{i=1}^n \left(\hat{R}^{\mathcal{M}}_{f}(x_i) - \hat{R}^{\mathcal{M}}_{(f, m_f)}(x_i)\right),
\end{equation}
where $\hat{R}^{\mathcal{S}}_{f^*}(x)$ is computed using the values of the $y_i$'s.

\subsection{Practical use}
To recap, two steps are required to compute the proposed evaluation metrics for a safety monitor in a specific application context. First, one needs to build/select a test dataset representative of the operational design domain and labeled with the correct values of $f^*$. Second, one need to design a procedure to compute $\hat{R}^{\mathcal{M}}_{(f,m_f)}(x)$ and $\hat{R}^{\mathcal{S}}_{(f,m_f)}(x)$, respectively representing the expected cumulative mission reward and safety reward from a given prediction and monitoring output. These choices are of utmost importance as they represent the underlying assumptions made about our system. Indeed, to guarantee the safety of an autonomous system using ML, one should not only ensure that it performs well under the chosen metrics, but also that the assumptions used to define the evaluation procedure are valid. As we will show in Section~\ref{sec:examples}, in most research dealing with ML monitoring, the latter is often disregarded. We believe that properly formulating these evaluation choices in the formalism proposed in this paper can help to understand the underlying assumptions made and, by extension, evaluate their validity. 

We emphasize that the proposed metrics are only valid to assess the performance of a monitor $m_f$ with respect to specific $\mathcal{D}$ and $f$. There is no guarantee that using the same monitoring approach with a different model and/or application context would result in similar performance. For safety-critical applications requiring monitoring, we do not believe that it is realistic to define application-agnostic and/or model-agnostic evaluation procedures. 

\section{Examples from the literature}\label{sec:examples}

This section presents three applications of runtime monitors to different types of perception functions. For each of these examples, we discuss how monitors have been evaluated previously, and we show how different evaluation approaches can be expressed in terms of our metrics. In particular, we show how to compute $\hat{R}^{\mathcal{S}}_{(f, m_f)}(x)$ and $\hat{R}^{\mathcal{M}}_{(f, m_f)}(x)$. By extension, $\hat{R}^{\mathcal{S}}_{f}(x)$ and $\hat{R}^{\mathcal{M}}_{f}(x)$ can be computed by setting $m_f$ to be the \textit{null} monitor (always returning zero), and $\hat{R}^{\mathcal{S}}_{f^*}(x)$ and $\hat{R}^{\mathcal{M}}_{f^*}(x)$ are computed by replacing $f(x)$ by its corresponding ground truth labels. For each use case, we also conduct practical experiments and compute the proposed metrics (SG, RH, AC) under different evaluation assumptions. We note that the objective of this section is not to design the most efficient monitors for each use case, but rather to demonstrate the use of the proposed metrics. Studying these applications through the prism of the formalism presented in Section~\ref{sec:approach} allows to highlight the underlying safety assumptions about the evaluated systems. The complete code to reproduce these experiments has been made publicly available\footnote{https://github.com/jorisguerin/MLSafetyMonitors-unifiedEvaluation-experiments}.

Before diving into the use cases, we remind the reader that the design of the evaluation dataset $D_{\text{eval}}$ is a crucial step to properly assess the performance of a perception system (Section~\ref{sec:approach_relax}). It represents the assumptions made about the actual operational design domain $\mathcal{D}$, i.e., the conditions that a system can encounter during execution. Here, we focus on demonstrating the usage of our metrics, and assume that the evaluation datasets used in our experiments are representative of their respective operational design domain, but in practice this claim would require strong justifications.

\subsection{Use Case 1 -- Classification}\label{sec:examples_classif}
Classification is the machine learning problem consisting in building a function $f$ mapping an input $x \in \mathcal{D}$ to an output $y \in [1, ..., K]$ among a predefined discrete set of $K$ categories. For classification, the generic goal of a runtime monitor is to identify ``bad input data''. Despite the simplicity of this definition, researchers have used different evaluation approaches to assess the performance of classification monitors. This section presents two popular evaluation schemes from the literature and shows how both can be expressed in terms of the proposed metrics. 
Preliminary experiments are proposed to compare these evaluations, and illustrate how changing the definition of the approximate return can lead to different results.

\subsubsection{Evaluation 1: Detect Model Errors}

For classification, the predictions of a model $f$ are either right or wrong (unlike other ML problems where correctness can be more gradual, e.g., object detection, semantic segmentation). This way, several works about runtime monitoring of classification functions have evaluated their monitors based on their ability to detect errors of $f$~\cite{neuron_activation_patterns, dissector, hendrycks2016baseline, benchmark_raul, doctor}. Hence, the monitor is viewed as a simple binary classifier and all traditional metrics from binary classifications (precision, recall, etc.) can be used to evaluate $m_f$ on $D_{\text{eval}}$. 

We now show how these traditional metrics can be expressed in terms of our proposed metrics. For classification monitors, we consider that missed detections decrease safety and wrong activations decrease availability. In our formalism, this translates by computing the \emph{safety return} for $x \in D_{\text{eval}}$ as
\begin{equation}\label{eq:returnHazard_classif}
\hat{R}^{\mathcal{S}}_{(f,m_f)}(x) = 
\begin{cases} 
    0 & \text{if } f(x) \neq y \text{ and }  m_f(x) = 0, \\
    1 & \text{else}.
\end{cases}
\end{equation}
And the \textit{mission return} as
\begin{equation}\label{eq:returnMission_classif}
\hat{R}^{\mathcal{M}}_{(f,m_f)}(x) = 
\begin{cases} 
    0 & \text{if } f(x) = y \text{ and }  m_f(x) = 1, \\
    1 & \text{else}.
\end{cases}
\end{equation}
In upcoming discussions, we consider that evaluation metrics are always computed on $D_{\text{eval}}$ and drop the exponents to simplify notations.

In the above definitions, the Safety Gain is only impacted by the true positives of $m_f$ (Equations~\ref{eq:SG_approx} and \ref{eq:returnHazard_classif}). If we divide $\hat{SG}_{m_f}$ by the fraction of wrong predictions of $f$, which is independent of $m_f$), the safety gain becomes the recall of the binary classifier $m_f$. Similarly, the Residual Hazard is only impacted by false negatives of $m_f$, and $\hat{RH}_{m_f}$ divided by the fraction of incorrect predictions of $f$ is the false negative rate of $m_f$. The Availability Cost only depends on false positives of $m_f$, and if we divide $\hat{AC}_{m_f}$ by the fraction of correct predictions of $f$, it becomes the false positive rate of $m_f$.

The main shortcoming of this evaluation scheme is that it does not allow to include additional knowledge about the system application context (e.g., from the hazard analysis). Indeed, Equations~\ref{eq:returnHazard_classif} and \ref{eq:returnMission_classif} only require information about the labeled evaluation data $D_{\text{eval}}$ and the associated predictions of $f$.
In other words, the estimates of the returns are independent of the system in which $f$ is integrated. This way, every misclassification of $f$ has an equal impact on both the safety and availability of the system. This assumption does not hold for many critical systems using classification, e.g. for an autonomous car road sign classifier, misclassifying a ''speed limit 30" as a ``speed limit 20" is not as severe as misclassifying a ''stop" as a ''speed limit 30". A tree-based approach for safety analysis of classification functions was proposed in~\cite{CFMEA}. It can be used within our formalism to extend this evaluation scheme and account for the asymmetry of the safety impact of different misclassifications.

\subsubsection{Evaluation 2: Detect Runtime Threats}
Instead of predicting model errors, other works have evaluated classification monitors on the surrogate problem of identifying specific kinds of runtime threats. A threat is defined as a change in the input data encountered at runtime, which can hinder model performance. In practice, researchers have tested their monitors on different types of threats such as Novelty~\cite{react, facer, lukina2020into} (label does not belong to any of the predefined classes), Distributional Shift~\cite{liang2018enhancing, hsu2020generalized} (image was not drawn from the training distribution, e.g., changes in external conditions, noisy sensors), Adversarial Attacks~\cite{kantaros2020visionguard, wang2019adversarial} (image was modified intentionally to deceive the monitored model). 

In this evaluation setup, the dataset is composed of normal images and threats. For an image $x \in D_{\text{eval}}$, a binary label $\tau$ represents whether it is a threat ($\tau = 1$) or not ($\tau = 0$). Then, one can define the \emph{safety return} for $x$ as
\begin{equation}\label{eq:returnHazard2_classif}
\hat{R}^{\mathcal{S}}_{(f,m_f)}(x) = 
\begin{cases} 
    0 & \text{if } \tau = 1 \text{ and }  m_f(x) = 0, \\
    1 & \text{else}.
\end{cases}
\end{equation}
And the \textit{mission return} as
\begin{equation}\label{eq:returnMission2_classif}
\hat{R}^{\mathcal{M}}_{(f,m_f)}(x) = 
\begin{cases} 
    0 & \text{if } \tau = 0 \text{ and }  m_f(x) = 1, \\
    1 & \text{else}.
\end{cases}
\end{equation}
Then, computing $\hat{SG}_{m_f}$, $\hat{RH}_{m_f}$ and $\hat{AC}_{m_f}$ is straightforward. The strong assumption under this setting is that the threat labels ($\tau$'s) represent the hazard associated with a prediction. In other words, we assume that $f$ can be trusted for in-distribution data ($\tau=0$) and that it should never be used under the identified threats ($\tau=1$). The second assumption is that the hazard associated with a prediction comes only from a specific characteristic of the input image and not the model being monitored. Indeed, in this setting, both the mission and safety returns are independent of $f$ (Eq.~\ref{eq:returnHazard2_classif} and \ref{eq:returnMission2_classif}). Hence, this evaluation setting seems unable to account for safety cases when $f$ does not present perfect accuracy on in-distribution data.

\subsubsection{Experiments}
To illustrate the influence of these evaluation choices, we conduct experiments on the CIFAR10 dataset~\cite{krizhevsky2009learning}. First, a neural network $f$ (3 convolutional layers followed by two dense layers) is fitted to the training set of CIFAR10, reaching a test accuracy of $0.79$. Then, to monitor this model, we consider two simple approaches:
\begin{enumerate*}
\item Features representing the training images are extracted from $f$, and several one-class classification (OCC) models are fitted independently to each class. In practice, we use features from the third layer of $f$ and an Isolation Forest~\cite{isolation_forest} for OCC. To foster conservativeness, the rejection threshold on the OCC scores is set such that $30\%$ of the training data are discarded. At runtime, features are extracted from new input images, and the OCC model decides whether to accept or reject them.
\item An auto-encoder is fitted to the CIFAR10 training set, such that the encoding part has the same structure as the convolutional block of the classifier. Then, the same monitoring approach is implemented using the features from the encoder.
\end{enumerate*} 

\begin{figure}[t]
     \centering
     \begin{subfigure}[b]{0.45\textwidth}
         \centering
         \includegraphics[width=\textwidth]{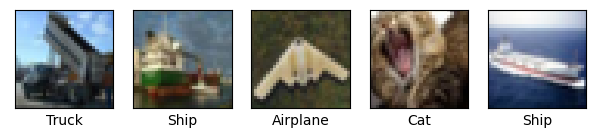}
         \caption{Images from the CIFAR10 test dataset.}
         \label{fig:cifar10}
     \end{subfigure}
     \hfill
     \begin{subfigure}[b]{0.45\textwidth}
         \centering
         \includegraphics[width=\textwidth]{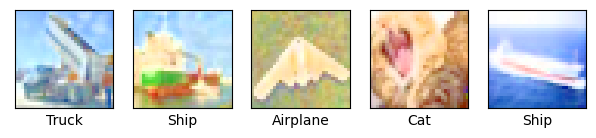}
         \caption{Corresponding images after brightness modification.}
         \label{fig:cifar10_bright}
     \end{subfigure}
     \hfill
\caption{Example images to illustrate the threat generation process used in our experiments.}
\label{fig:expe_classif}
\end{figure}

For evaluation, we generated distributional shift threats by modifying lighting conditions of the ten thousand test images of CIFAR10 (Figure~\ref{fig:expe_classif}). The proposed metrics are computed using both evaluation schemes presented above: predicting model errors (E1)  and detecting runtime threats (E2). E1 evaluates the ability of a monitor to recover errors of the neural network, while E2 evaluates its ability to recover images with modified brightness.

The results obtained are presented in Table~\ref{tab:results_classif}. We can see that the choice of the approximate return functions influences greatly the values of the proposed metrics. For example, under E1, the auto-encoder monitor has a higher Availability Cost than the Classifier monitor. This makes sense as the auto-encoder is independent of $f$ and does not contain information regarding its predictions. On the other hand, under E2, the auto-encoder monitor has a much higher Safety Gain than its classifier counterpart, which means that auto-encoder features are better at detecting runtime images with modified brightness. 

\begin{table}[t]
    \centering
    \caption{Results of our classification experiments.}
    \begin{tabular}{c|ccc|ccc}
     & \multicolumn{3}{c}{Auto-encoder} & \multicolumn{3}{|c}{Classifier} \\
         & SG & RH & AC & SG & RH & AC  \\ \hline
    E1 & 0.184 & 0.140 & 0.304 & 0.074 & 0.251 & 0.154 \\
    E2 & 0.344 & 0.156 & 0.144 & 0.086 & 0.414 & 0.142
    \end{tabular}
    \label{tab:results_classif}
\end{table}

This example highlights the importance of defining evaluation goals ($\hat{R}^{\mathcal{M}}$ and $\hat{R}^{\mathcal{S}}$) aligned with the high-level objectives of the monitored system. Indeed, by studying two popular evaluation schemes from the literature, we showed how these choices could alter our perception of the safety and availability of a given monitor. We do not claim here that one evaluation scheme is better than the other. However, we believe that following our formalism to evaluate a runtime monitor is a good way to ensure compliance with the application objectives. At the same time, the proposed metrics allowed us to gain a unified and interpretable insight regarding the monitor performance.

\subsection{Use Case 2 -- Object Detection for Pedestrian Avoidance}

\begin{figure}[t]
    \centering
    \begin{subfigure}{0.22\textwidth}
    \includegraphics[width=\textwidth]{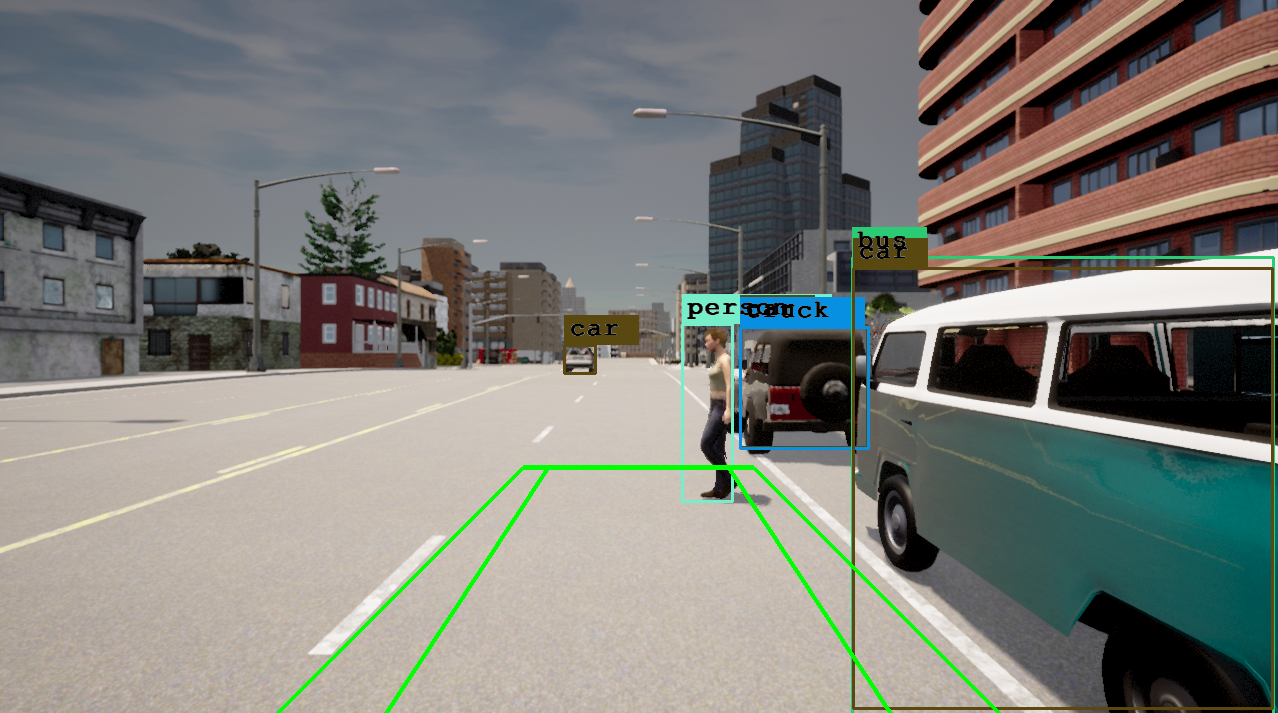}
    \caption{No perturbation. No pedestrian detection error.}\label{fig:carla_normal}
    \end{subfigure}
    ~
    \begin{subfigure}{0.22\textwidth}
    \includegraphics[width=\textwidth]{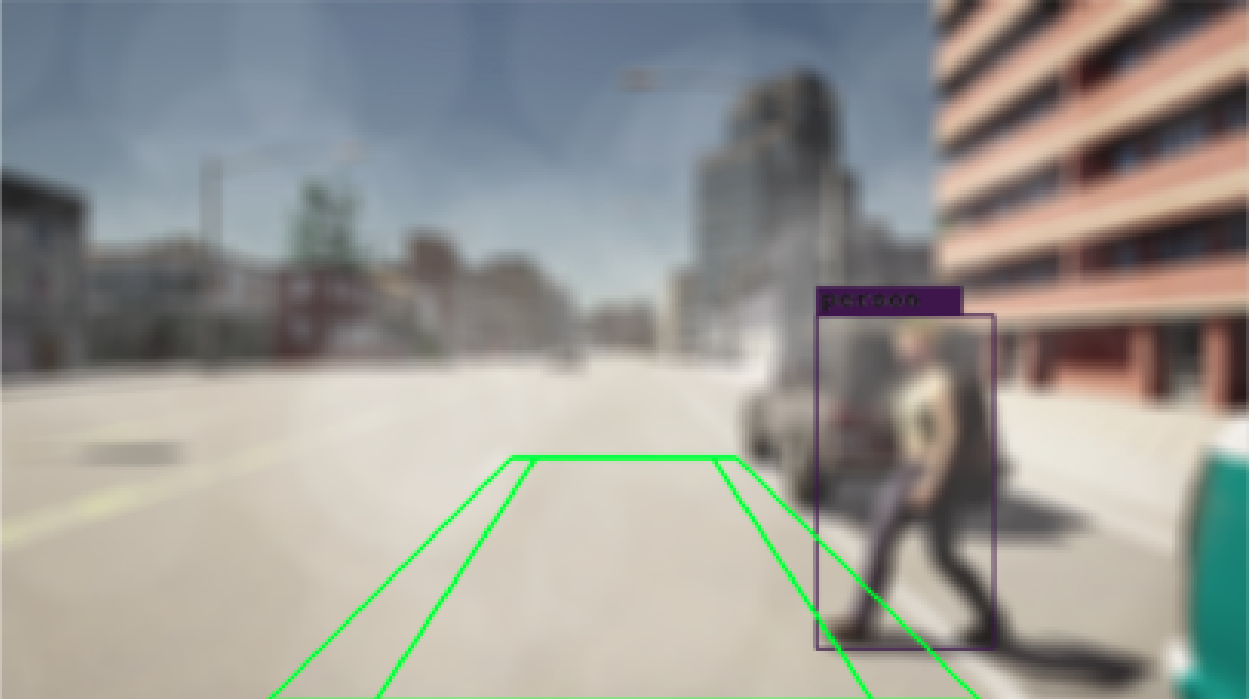}
    \caption{Smoke. No pedestrian detection error.}\label{fig:carla_fog}
    \end{subfigure}
    
    \begin{subfigure}{0.22\textwidth}
    \includegraphics[width=\textwidth]{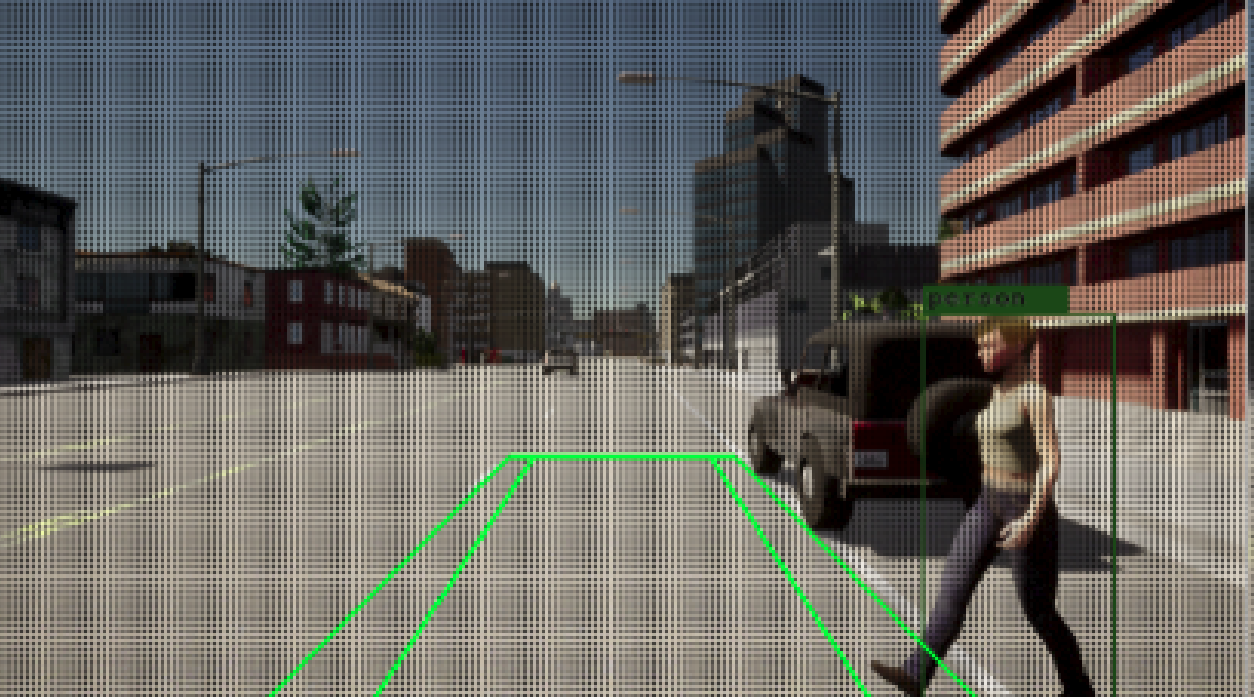}
    \caption{Grid dropout. No pedestrian detection error.}\label{fig:carla_grid}
    \end{subfigure}
    ~
    \begin{subfigure}{0.22\textwidth}
    \includegraphics[width=\textwidth]{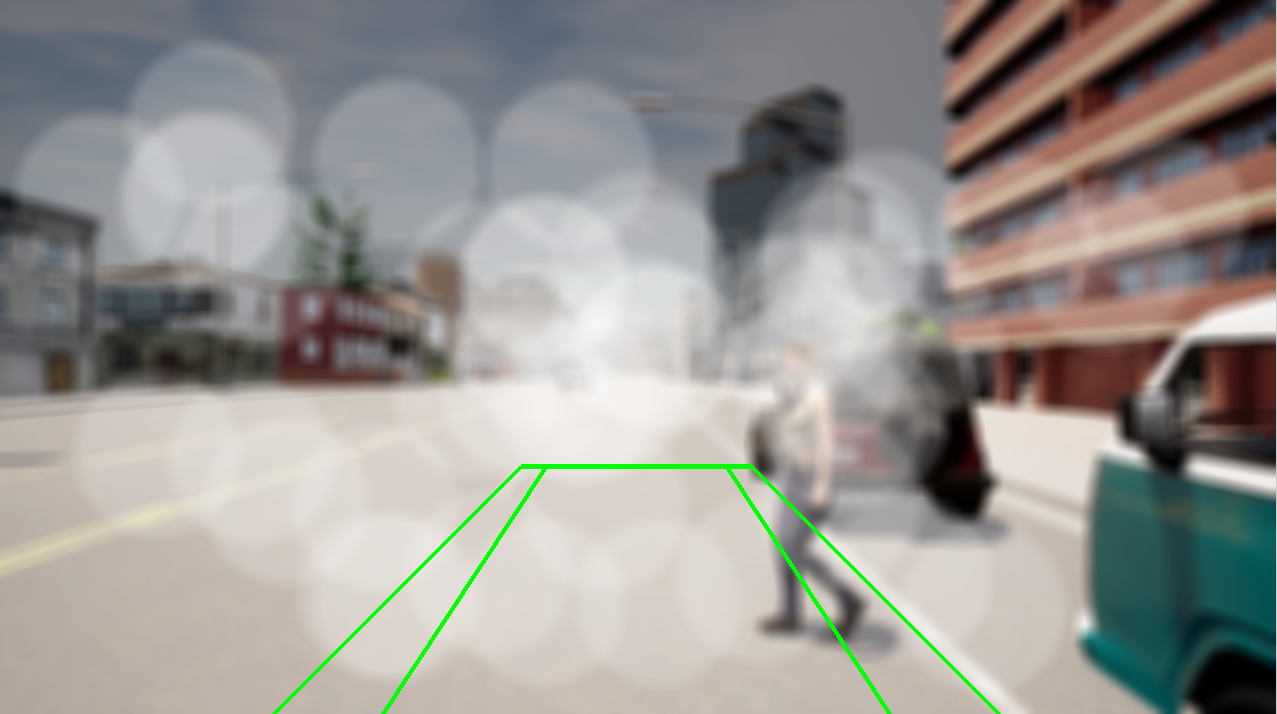}
    \caption{Sun flare. False Negative of the pedestrian detector.}\label{fig:carla_sun}
    \end{subfigure}
    \caption{\textbf{Pedestrian avoidance use case}. Each image represents a different perturbation.}
    \label{fig:carla}
\end{figure}

Our second example is an autonomous driving scenario that we designed on the CARLA simulator~\cite{Dosovitskiy17}. A car is equipped with an object detection model (YOLO-v5~\cite{glenn_jocher_2021_5563715} trained on COCO~\cite{lin2014microsoft}), which is used to locate pedestrians within an emergency braking system. The scenario is designed as follows: the car drives normally in a city road, and after a few seconds, a pedestrian appears suddenly in front of the vehicle after crossing between parked cars (Figure~\ref{fig:carla}). The emergency braking system receives information from the pedestrian detector, and stops the vehicle whenever a pedestrian bounding box overlaps with a predefined safety-critical region (green region in Figure~\ref{fig:carla}). The scenario presented here runs for a fixed number of steps (episodes of $T$ frames). We also add a runtime monitor to this system, consisting of two modules: \begin{enumerate}
\item The contrast variation in the input images is monitored, and if it falls below a predefined threshold, the monitor is activated. The objective of this module is to detect faulty images due to harsh environmental conditions (e.g., fog), or sensor failures (e.g., blur).
\item A plausibility checker ensures that bounding boxes in consecutive frames are consistent, e.g., no sudden appearance of bounding boxes in front of the vehicle. Such inconsistencies can indicate ghost detections. 
\end{enumerate}
When the monitor detects an anomaly, the emergency breaking system immediately stops the vehicle. For this use case as well, we propose two different ways of computing the safety and mission rewards, and compare their influence on the proposed evaluation metrics.


\subsubsection{Evaluation 1: Detect ML Errors}
This evaluation scheme is the adaptation of Section~\ref{sec:examples_classif} for object detection. To identify errors of an object detector, one must compare ground truth bounding boxes to bounding boxes predicted by the model. To boxes are considered to match if their labels are identical, and both the prediction score and their intersection over union are above fixed thresholds. The binary variable $\tau$ represents the error status of the monitored object detector for an image $x$. We consider that there is an error ($\tau=1$) if there is either a false positive (prediction without corresponding ground truth) or a false negative (ground truth without prediction). Then, the safety and mission returns are
\begin{equation}\label{eq:returnHazard2_AV}
\hat{R}^{\mathcal{S}}_{(f,m_f)}(x) = 
\begin{cases} 
    0 & \text{if } \tau = 1 \text{ and }  m_f(x) = 0, \\
    1 & \text{else}.
\end{cases}
\end{equation}
\begin{equation}\label{eq:returnMission2_AV}
\hat{R}^{\mathcal{M}}_{(f,m_f)}(x) = 
\begin{cases} 
    0 & \text{if } \tau = 0 \text{ and }  m_f(x) = 1, \\
    1 & \text{else}.
\end{cases}
\end{equation}

\subsubsection{Evaluation 2: Prevent accidents}

This second approach aims to assess the ability of a monitor to prevent accidents. Conducting evaluations in a simulation environment such as Carla presents several advantages. First, we are able to know the exact location of every object of interest at each time step, which allows to define the mission and safety rewards in terms of the exact world state. Second, any initial conditions can be reproduced to test the system's behavior with different perception function (e.g., with or without the monitor). Hence, the proposed evaluation metrics can be defined directly from the values of reward functions at individual time steps. 

The mission reward $r^{\mathcal{M}}(z_t)$, associated with configuration $z_t$ is defined to be 0 when the car is stopped and 1 when it is running. In our experiments, we make the simplifying assumption that, when requested, the emergency braking system stops the vehicle instantly (from one frame to the next). Hence, the predictions obtained at time step $t$ only impact the mission reward at $t+1$, and to compute the mission return, Equation~\ref{eq:defReturnMission} can be adapted as:
\begin{equation}\label{eq:defReturnMission_AV}
    \hat{R}^{\mathcal{M}}_{(f,m_f)}(x) = r^{\mathcal{M}}(z_{t+1}|(f, m_f)(x_t)).
\end{equation}
Finally, we also consider that when emergency braking is triggered, the episode ends and the mission rewards for all consecutive steps is $0$. This way, for a given episode, the availability cost is the average difference in the number of running frames with and without the monitor.  

The safety reward ${r}^{\mathcal{S}}(z_t)$ is defined slightly differently. It is -1 for frames when there is a collision with the pedestrian and 0 for all other frames. Hence, the safety return is: 
\begin{equation}\label{eq:defReturnSafety_AV}
    R^{\mathcal{S}}_{(f, m_f)}(x_t) = \sum_{k=t+1}^{T} r^{\mathcal{S}}(z_k|(f, m_f)(x_t)),
\end{equation}
In other words, a prediction $(f, m_f)(x_t)$ gets a negative safety reward if there is an accident in its future ($\{z_{t+1}|(f,m_f)(x_t), ..., z_{T}|(f,m_f)(x_t)\}$).

\subsubsection{Experiments}

Our experimental evaluation consist in running the same simulation scenario several times, while injecting different kinds of faults. In particular, we generate twelve types of image perturbation presented in~\cite{secci2020failures,buslaev2020albumentations,benchmark_raul}: smoke, sun flare, rain, row add logic, shifted pixel, coarse dropout, grid dropout, channel shuffle, channel dropout, contrast, brightness, and Gaussian noise (see Figure~\ref{fig:carla} for examples). For most perturbations, several intensity levels are tested, resulting in a total of 53 simulations. The fixed number of time steps is set to $T = 220$.

The proposed metrics are computed using both evaluation schemes: detect ML errors (E1) and prevent accidents (E2). The results are presented in Table~\ref{tab:results_AV}. Using E1, it seem that the monitor allows to handle most of the hazards present in the system (high SG and low RH) and does not impact much model performance (low AC). However, when the true temporal behavior of the system is considered (E2), the results appear much less enthusiastic.
Our experiments included $15\%$ of scenarios with an accident (SG + RH), and the monitor allowed to avoid only half of them. In the meantime, the availability was decreased by $80\%$.

\begin{table}[t]
    \centering
    \caption{Results of our pedestrian avoidance experiments.}
    \begin{tabular}{c|ccc}
         & SG & RH & AC \\ \hline
    E1 & 0.187 & 0.060 & 0.065 \\
    E2 & 0.075 & 0.075 & 0.800 
    \end{tabular}
    \label{tab:results_AV}
\end{table}


\subsection{Use Case 3 -- Semantic Segmentation for UAV Emergency Landing}\label{sec:examples_EL}

Our last use case is built on recent work about emergency landing of Unmanned Aerial Vehicles (UAV) in urban environments~\cite{icra22}. This module is triggered when the UAV loses its localization capabilities. Then, it collects an onboard image, reduces its resolution, and process it with a semantic segmentation model $f$ that classifies each pixel in one of the following categories: building, road, car, tree, low vegetation, humans, and background. Following the safety analysis conducted in~\cite{ssiv21}, the binary safety flag $\varphi_k$ is defined to represent when a category $C_k$ is too dangerous to land ($\varphi_k = 1$). Here, the goal is to avoid roads and buildings. In addition, a hazard score $h_k \in [0,1]$ is defined for all accepted categories ($\varphi_k = 0$). 

\begin{figure}[t]
    \centering
    \includegraphics[width=0.45\textwidth]{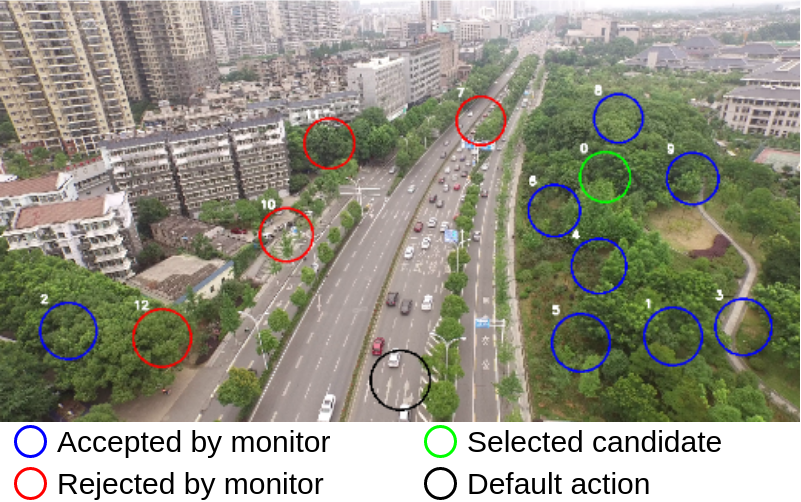}
    \caption{\textbf{Example outputs for Emergency Landing}. Circles are candidates identified by the main model, colors indicate runtime monitor status. The default action consists in opening a parachute without moving. Source: \cite{icra22}.}
    \label{fig:EL}
\end{figure}

A candidate landing area is a circular region of fixed size radius, containing only safe pixels (Figure~\ref{fig:EL}). The candidate with the lowest hazard score ($\sum h_k$) is selected for landing.  To increase confidence in $f$, we use the ``local high definition'' monitor. It consists in using the full resolution image to refine the semantic segmentation of small patches containing the candidates, allowing to improve critical predictions, while still controlling computation time. The monitor rejects a candidate if its new segmentation contains a forbidden pixel ($\varphi_k = 1$). If no suitable candidate was found by the $(f, m_f)$ pair, the default action consists in stopping the motors and opening a parachute at the current location. Figure~\ref{fig:EL} illustrates the outputs of emergency landing on an example image.

\subsubsection{Evaluation 1 -- Detect wrong candidates} As its name implies, emergency landing is an emergency procedure which only goal is to ensure people's safety. Hence, for this application, we are not concerned with mission progress and availability cost is not a relevant metric. The safety return of a candidate can be computed using Equation~\ref{eq:returnHazard2_AV}, where $\tau=1$ for candidates containing unsafe pixels. Then, the safety return for the entire image is simply the average across candidates.

\subsubsection{Evaluation 2 -- Compare selected landing zones}
Another way of evaluating a monitor is to compare the final candidate that was actually selected for landing with and without it. 
Let $(x, y)$ be an input image and its associated ground truth segmentation image, and let $y(p)$ be the ground truth class of pixel $p$. The safety score of a landing candidate $\mathcal{L}_c$ in $x$ is
\begin{equation}
S(x[\mathcal{L}_c]) = 
\begin{cases}
    0 \;\;\;\;\;\;\;\;\, \text{if } \exists p \in x[\mathcal{L}_c] \; s.t. \; \varphi_{y(p)} = 1, \\
    \kappa + (1-\kappa) \mathbb{E}_p [1-h_{y(p)}] \;\;\;\;\;\;\;\;\; \text{else},
\end{cases}
\end{equation}
where $\kappa \in [0,1]$ defines a gap to separate unsafe candidates from others. From this definition, $\hat{R}^{\mathcal{S}}_{(f,m_f)}(x)$ can be defined as the value of $S(x[\mathcal{L}_c])$ when $(f, m_f)(x)$ selects landing zone $\mathcal{L}_c$.

\subsubsection{Experiments}
The proposed experiments are conducted on the 70 images of the validation set of UAVid~\cite{uavid}. Five types of perturbations are applied (brightness, fog, motion blur, pixel trap and shifted pixels), leading to a total of 420 images. Both evaluation schemes are compared to evaluate the performance of the proposed monitor on the emergency landing application. 

\begin{table}[t]
    \centering
    \caption{Results of our emergency landing experiments.}
    \begin{tabular}{c|cc}
         & SG & RH \\ \hline
    E1 & 0.805 & 0.195 \\
    E2 & 0.108 & 0.212 
    \end{tabular}
    \label{tab:results_EL}
\end{table}

Results are presented in Table~\ref{tab:results_EL}. Once again, we observe that evaluating the monitor as a regular binary classifier (E1) is much more optimistic than considering its performance at system level (E2). For this specific example, this discrepancy comes from the fact that rejecting a valid landing zone can severely impact safety, as other safe options might not exist. This subtlety of emergency landing is well captured by E2, but not by E1, which ignores false positives. We also note that, using E2, the safety gain can sometimes be negative when the monitor rejects good options.

\section{Conclusion}\label{sec:conclusion}
This paper presents new evaluation metrics for runtime monitoring of ML perception, called \emph{Safety Gain} (SG), \emph{Residual Hazard} (RH), and \emph{Availability Cost} (AC). These metrics represent different aspects of the system in which the ML function is used: the safety benefits of using the monitor (SG), the remaining threats despite the monitor (RH), and the negative impact of the monitor on the system's performance (AC). This formulation relies on expressing the future returns (cumulative rewards) for safety and mission objectives. To show that these metrics are generic and flexible, we demonstrated how they can be used for three independent use cases, representing different application contexts. Experiments on each use case demonstrate the importance of properly formulating the assumptions about the system evaluated. Indeed, we show that different definition of the return functions can drastically change the perceived performance of the monitor. In summary, this work is a step towards unifying the field of ML runtime monitoring, allowing to compare approaches across different scenarios using common criteria. Using our formalism helps to ensure that evaluation is aligned with the actual system requirements.

A possible future work direction would consist in building a set of evaluation scenarios representing real-world applications of ML, using both simulated environments and real-world data. This would allow a proper benchmark study of existing runtime monitoring techniques using the metrics introduced in this paper. Such natural extensions of our work has the potential to play a major role in the development of future autonomous systems, as it allows for a better assessment of the safety of such critical ML-based functions.


\section*{Acknowledgements}
This research has benefited from the AI Interdisciplinary Institute ANITI. ANITI is funded by the French ”Investing for the Future – PIA3” program
under the Grant agreement No ANR-19-PI3A-0004.

This research has also received funding from the European Union’s Horizon 2020 research and innovation program under the Marie Skłodowska-Curie grant agreement No 812.788 (MSCA-ETN SAS). This publication reflects only the authors’ view, exempting the European Union from any liability. Project website: http://etn-sas.eu/.

\end{document}